\documentclass[10pt,journal,compsoc]{IEEEtran}
\ifCLASSOPTIONcompsoc
  \usepackage[nocompress]{cite}
\else
  \usepackage{cite}
\fi
\ifCLASSINFOpdf
   \usepackage[pdftex]{graphicx}
  \graphicspath{{images/}}
\else
\fi
\usepackage{amsmath}
\usepackage{amsfonts}
\usepackage{multirow}

\begin{document}
%
\title{Multiview Deep Learning for Predicting Twitter Users' Location}
%
%
%
%

\author{{Tien~Huu~Do,
        Duc~Minh§~Nguyen,~
        Evaggelia~Tsiligianni,~
        Bruno Cornelis,~\\
        and~Nikos~Deligiannis,~\IEEEmembership{Member,~IEEE}
\thanks{
T.~H.~Do, D.~M.~Nguyen, E.~Tsiligianni, B. Cornelis and N.~Deligiannis are with the Department of Electronics and Informatics (ETRO), Vrije Universiteit Brussel, Brussels B-1050, Belgium, and also with IMEC, \mbox{Leuven} B-3001, Belgium (e-mail: thdo@etrovub.be, mdnguyen@etrovub.be, \mbox{etsiligi@etrovub.be}, bcorneli@etrovub.be, ndeligia@etrovub.be).
}
}
}

\IEEEtitleabstractindextext{%
\begin{abstract}
The problem of predicting the location of users on large social networks like Twitter has emerged from real-life  applications such as social unrest detection and online marketing. Twitter user geolocation is a difficult and active research topic with a vast literature. Most of the proposed methods follow either a content-based or a network-based approach. The former exploits user-generated content while the latter utilizes the connection or interaction between Twitter users. In this paper, we introduce a novel method combining the strength of both approaches. Concretely, we propose a multi-entry neural network architecture named MENET leveraging the advances in deep learning and multiview learning. The generalizability of MENET enables the integration of multiple data representations. In the context of Twitter user geolocation, we realize MENET with textual, network, and metadata features. Considering the natural distribution of Twitter users across the concerned geographical area, we subdivide the surface of the earth into multi-scale cells and train MENET with the labels of the cells. 
We show that our method outperforms the state of the art by a large margin on three benchmark datasets.
\end{abstract}

\begin{IEEEkeywords}
Twitter user geolocation, deep learning, feature learning, multiview learning, big data.
\end{IEEEkeywords}}

\maketitle

\IEEEdisplaynontitleabstractindextext

%
\IEEEpeerreviewmaketitle

\section{Introduction}

Over the last ten years, social networks have grown and engaged a massive amount of users. Among them, Twitter is one of the most popular, reaching over 300 million users by the $4^{th}$ quarter of  $2017$~\cite{twitter2017statista}. On Twitter, users publish short messages of 140 characters or less called \textit{tweets}, which can be seen by followers or by the public. 
Tweets can also be re-published by users who have seen the tweets, a process known as \textit{retweeting}. 
This way, information can be spread quickly and widely throughout the whole Twitter network. Twitter can even be considered as a human-powered sensing network, with a lot of useful information, yet, in an unstructured form. For this reason, automatic mining and extracting meaningful information from the massive amount of Twitter data is of great significance~\cite{minab2014online},~\cite{ sechelea2016twitter}.

A very useful piece of information on Twitter is user location, which enables several applications including event detection~\cite{sakaki2013tweet}, online community analysis~\cite{Komorowski2017}, social unrest forecasting~\cite{compton2013detecting} and location-based recommendation~\cite{bao2015recommendations,zhao2017service}. As another example, user location information can be useful for online marketers and governments to understand trends and patterns ranging from customer and citizen feedback~\cite{celikten2017modeling} to the mapping of epidemics in concerned geographical areas~\cite{ji2013monitoring}. In 2009, Twitter enabled a geo-tagging feature, with which users can choose to geo-tag their tweets while posting. However, the majority of tweets are not geo-tagged by the users~\cite{ICWSM1510561}. Alternatively, users' location might be available via their profile  data. Nonetheless, not many users disclose their location via their Twitter profile, or the provided information is often unreliable. For example, a user might share vague or non-existent places such as "Everywhere" and "Small town, RW Texas". This results in a quest for geolocation algorithms that can automatically analyze and infer  the location of Twitter users. 

The Twitter geolocation problem can be addressed at two different levels, namely, the tweet level and the user level. The former aims at predicting the location of single tweets, while the latter aims at inferring the location of a user  from the data generated by that user. 
The geolocation of single tweets is extremely difficult due to the limited availability of information.  Research on single tweet geolocation has been conducted~\cite{duong2016near,lau2017end}, but a good accuracy can be achieved only under specific constraints, which are normally not applicable in real-life situations. On the other hand, the Twitter geolocation at user level, also refered to as \textit{Twitter user geolocation}, is more common, with plenty of methods described in the literature~\cite{jurgens2015geolocation}. In this paper, we focus on the geolocation problem at user level instead of tweet level. 

The Twitter user geolocation problem can be formulated under a classification or a regression setting. Under the classification setting, one can predict the location of users in terms of geographical regions, such as countries, states and cities. Under the regression setting, the task is to estimate the exact geocoordinates of the users. Both prediction settings are considered in this paper. It is worth mentioning that we address the regression problem from a classification point of view. Towards this end, we employ a map partitioning technique to divide the concerned geographical area into small regions corresponding to classes. The exact geocoordinates of Twitter users can be estimated using the classes' centroids.

In the Twitter user geolocation literature, most of the existing algorithms follow either a content-based approach or a network-based approach. Content-based methods extract information from the  textual contents of tweets to predict user locations~\cite{ICWSM1510561,LiuPaper,priedhorsky2014inferring}. Network-based methods, on the other hand, employ connections between users for geolocation~\cite{backstrom2010find,ICWSM136067,compton2014geotagging}. Both approaches have achieved good geolocation accuracy~\cite{ICWSM1510561,rahimi2017}.

This paper explores a more generic approach, which inherits the advantages of both content-based and network-based strategies.
Our approach leverages recent advances in deep neural networks (i.e., deep learning) and multiview learning.
Deep neural networks~\cite{bengio2015deep}, have been proven to be very effective in many domains including image classification~\cite{krizhevsky2012imagenet}, machine translation~\cite{luong2014addressing},  and compressive sensing~\cite{nguyen17}.
On the other hand, multiview learning is an emerging paradigm encompassing methods that learn from examples with multiple representations~\cite{ZHAO201743} showing a great progress recently~\cite{zhang2012combining,yu2012combining}. In Twitter user geolocation, the views can be different types of information available on Twitter such as text and metadata, or even features  extracted from the tweets themselves. 

Our contributions in this work are as follows:
\begin{itemize}
   \item We propose a generic multiview neural network architecture, named multi-entry neural network (MENET), for Twitter user geolocation.
   MENET is capable of combining multiview features into a unified model to infer users' location.
   \item We propose to incorporate four specific types of features to realize MENET for  Twitter user geolocation. These features capture the textual information (\textit{TF-IDF}, \textit{doc2vec}~\cite{quocle2014}), the user interaction network structure (\textit{node2vec}~\cite{grover2016node2vec}), and the time-related user behavior. 
   \item We show the effectiveness of using map partitioning techniques in Twitter user geolocation, especially with Google's S2 partitioning library\footnote{https://code.google.com/archive/p/s2-geometry-library/}. We have achieved state-of-the-art results on several popular datasets with these partitioning techniques.
   \item We show a thorough analysis on the importance of input features and the impact of partitioning strategies on the performance of MENET.
\end{itemize}

The remainder of this paper is organized as follows. In Section~\ref{related_work}, we review related works. Section~\ref{methodology} describes our method in details, including the model architecture, feature learning, feature extraction and how we improve our model with the density-driven map partitioning technique. Section~\ref{experiments} describes the performance criteria, the pre-processing procedures and details the parameter setting of our method. The results of our experiments are also presented in this section. Finally, we draw the conclusion and discuss future work in Section~\ref{conclusion_future_work}.

\section{Related Work}\label{related_work}
Most current approaches for predicting the location of Twitter users are based either on user-generated content or on the social ties. The first approach, which has been investigated thoroughly, uses textual features from tweets to build location predictive models. The latter arises from an observation that a user often interacts with people in nearby areas~\cite{backstrom2010find}, and exploits the network connections of users. This section will bring a closer look on recently published works for both approaches.

Plenty of content-based methods have been proposed for Twitter user geolocation. Geographical topic models~\cite{hong2012discovering,eisenstein2010} consider tweets and locations as the outputs of a generative process incorporating topics and regions as latent variables, thus geo-locating users by seeking to recover these variables. An alternative approach is using geographical Gaussian Mixture Models (GMMs)~\cite{priedhorsky2014inferring} to model the distribution of terms of tweets across geographical areas. By calculating a weighted sum of corresponding GMMs on terms of tweets, a geographical density function can be found, revealing the location at the single tweet level. A smilar approach, making use of GMMs, is introduced by Chang~\textit{et al.}~\cite{chang2012}, where a GMM model is fit to the conditional probability of a certain city, given a term. Char \textit{et al.}~\cite{ICWSM1510561} estimate location by exploiting the expressiveness of sparse coding and the advances in dictionary learning to obtain the state of the art on a benchmark dataset named GeoText~\cite{eisenstein2010}. Recently, several methods have addressed the Twitter user geolocation problem using deep learning. For example, Liu and Inkpen train stacked denoising autoencoders for predicting regions, states, and geographical coordinates~\cite{LiuPaper}. These vanilla models obtain quite good results with a pre-training procedure. 
These methods, however, do not take into account the natural distribution of Twitter users in the considered datasets over the different regions of interest. Concretely, the density of Twitter users is much higher in inner-city areas than countrysides. To exploit this attribute, grid-based geolocation methods are introduced in~\cite{roller2012,wing2011geodesic,wing2014hierarchical,melo2015geocoding}, where adaptive or uniform grids are created to partition the datasets into geographical cells at different levels. The prediction of geographical coordinates is then converted to a classification problem using the cells as classes, and off-the-shelf classifiers can be applied directly. This strategy is also used in our method but with a different spliting scheme and with a novel model architecture.

Recent works have shown a correlation between the likelihood of friendship of two social network users and the geographical distance between them~\cite{backstrom2010find}. Using this correlation, the location of users can be estimated using their friends' location. This is the key idea behind the network-based approach. By leveraging the social interactions like bi-directional following\footnote{Twitter users follow other people to see their latest updates. Bi-directional following means two users follow each other.} and bi-directional mentioning\footnote{Twitter users can mention other people in their tweets by typing \texttt{@username}. Bi-directional mentioning is the two-way interaction which happens when two users have mentioned each other.}, one can establish graphs of Twitter users where a label propagation algorithm~\cite{raghavan2007near} or its variants~\cite{baluja2008video,talukdar2009new} are used to identify locations of unlabeled users~\cite{davis2011,ICWSM136067,compton2014geotagging,apreleva2015predicting}. The network-based approach has several advantages over the content-based counterpart, including language independence. Also, it does not require training, which is a very resource intensive and time-consuming process on big datasets.
However, the inherent weakness of this approach is that it cannot propagate labels (locations) to users that are not connected to the graph. As a result, isolated users remain unlabeled.

To address the problem of isolated users in the network-based approach, unified text and network methods are proposed in~\cite{rahimi2015,rahimi2017}, which leverage  both the discriminative power of textual information and the representativeness of the users' graph. In particular, the textual information is used to predict labels for disconnected users before running label propagation algorithms. Additionally, the novelty of the works~\cite{rahimi2015,rahimi2017} lies in building a densely undirected graph based on the mentioning of users. This makes a significant improvement in the location prediction. 
Following~\cite{rahimi2015,rahimi2017}, models combining text, metadata and user network features have been introduced~\cite{miura2017unifying,lau2017end}. These models have to rely on user profile information including user location, user timezone and user UTC offset. These types of information should be considered unvailable in the Twitter user geolocation context. 
That is the reason why the three benchmark datasets considered in this paper do not provide the Twitter profile information.


Our method does not rely on the Twitter user profile information. 
It employs a similar graph of Twitter users derived from tweets as in~\cite{rahimi2015}; however, instead of propagating labels through the graph, our method trains an embedding mapping function to capture the graph's structure. The graph feature is then integrated with all other features in a neural network architecture. Our architecture is simpler as it does not require designing a specific architecture for each type of feature like in~\cite{miura2017unifying,lau2017end}, thus easier and less resource intensive to train.

\section{Multi-entry Neural Network for Twitter User Geolocation}\label{methodology}

\begin{figure*}[t]
\centering
\includegraphics[width=0.6\textwidth]{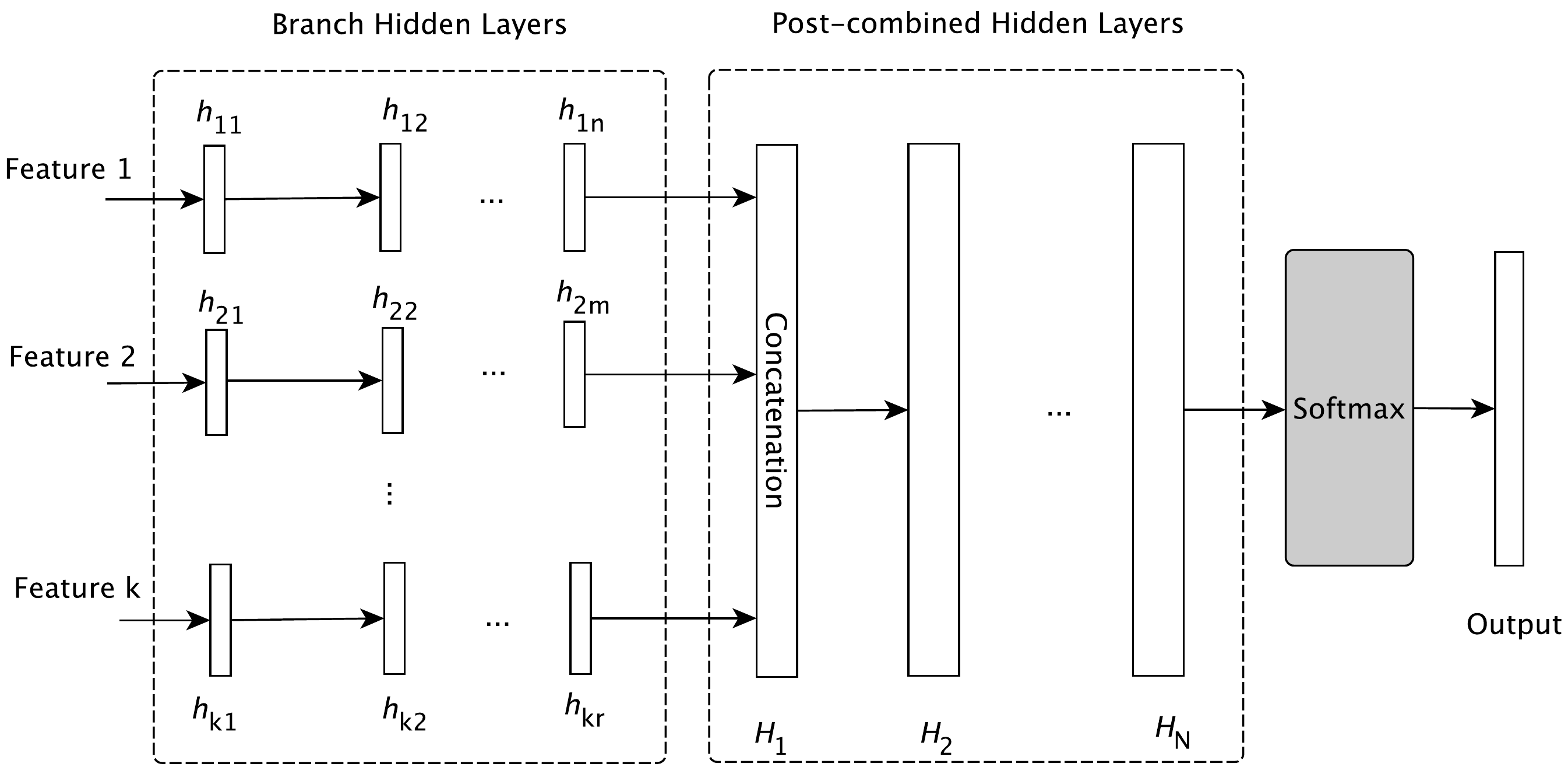}
\caption{Generic architecture for multi-entry neural network (MENET). MENET accepts various features to its input branches. Each branch may contain many hidden fully connected layers. Also, more fully connected layers can be added after the concatenation layer.}
\label{fig:model1}
\end{figure*}

In Twitter user geolocation, we wish to predict the location of a user using textual information and metadata, obtained from a corpus of tweets sent by the user, as well as information extracted from the user's network. Using this information, we predict either the area (alias, region), where the user most probably resides, or even the location of the user by means of geocoordinates.
Our method addresses this problem as a classification problem. Concretely, for each considered dataset, we subdivide Twitter users into discrete geographical regions, which correspond to classes. We define the centroid of a region by the median value of the geocoordinates of all training users in that region. Once a test user is classified to a certain region, we consider the centroid of that region as the predicted geocoordinates.

We propose a generic neural network model to learn from multiple views of data for Twitter user geolocation. 
We coin the proposed model MENET. 
The advantage of this model is the capability of exploiting both content-based and network-based features, as well as other available features concurrently. 
In this work, we realize MENET with different types of features. These features capture not only the tweets' content, but also the user network structure and time information. 
It is worth mentioning that except the time information, all other features are extracted from the tweets' content. Hence, MENET works even in case tweets' metadata is not available.
Integrating all features into MENET results in a powerful method for geolocation. 
Combining this method with the Google S2 map partitioning technique, we achieve state-of-the-art results in several Twitter user geolocation benchmarks.  
This section presents our MENET model and the different types of employed features in detail. 

\subsection{Model Architecture} 
\label{Model_Architecture}
\subsubsection{Architecture}

Our MENET architecture is illustrated in Fig.~\ref{fig:model1}. 
The model leverages different features extracted from the tweets' content and metadata. 
Each corresponds to one view of the network. 
In Fig.~\ref{fig:model1}, $k$ features are put into $k$ individual branches. Each branch can contain multiple hidden layers allowing to learn higher order features. 

Given multiple views of the input data, a straightforward approach to combine them is to use vector concatenation. 
Nevertheless, we argue that our architecture is more effective. Simple vector concatenation often does not fully utilize the power of multiple features. In MENET,
each view is the input to one network branch, which comprises of a number of fully connected hidden layers. 
In order to learn a non-linear transformation function for each branch, we employ the ReLU~\cite{article:relu} activation function after each hidden layer. 
The ReLU function is efficient for backpropagation and less prone to the vanishing gradient problem~\cite{bengio1994learning} than the \emph{tanh} and \emph{sigmoid} activation functions, hence, 
has been used widely in deep learning literature~\cite{maas2013rectifier,pan2016expressiveness}. The outputs of these branches are concatenated making a combined hidden layer. More fully connected layers can be added after this concatenation layer to gain more nonlinearity (see component \textit{Post-combined Hidden Layers} in Fig.~\ref{fig:model1}). Again, ReLU is used to activate these layers. At the end, we employ a softmax layer~\cite{Softmax} to obtain the output probabilities.


We employ the cross-entropy loss as the objective function. Let $N$ be the number of examples and $m$ be the number of classes, then the cross-entropy loss is defined by:
\begin{equation} \label{eq:cross_entropy}
L = -\sum_{i=1}^{N} \sum_{j=1}^{m} y_{i}^{j} \log(\tilde{y}_i^{j}), 
\end{equation}
where $y_{i}$, $i=1,\dots , N$ is the ground-truth vector, $\tilde{y}_i$ is the predicted  probability vector, namely, $\tilde{y}_i^j$ is the probability that user $i$ resides in region $j$.

\subsubsection{Training MENET}\label{overfitting_handling}

We train MENET using the stochastic gradient descent (SGD) algorithm~\cite{bottou2012stochastic}, which optimizes the objective function in~\eqref{eq:cross_entropy}.
In order to avoid overfitting, we use $\ell_2$ regularization and early stopping techniques. 
The $\ell_2$ regularization adds an additional term to the objective function, penalizing weights with big absolute values. 
Even though it is common practice to regularize weights in all layers, we empirically found that regularizing only 
the final output layer still effectively avoids overfitting, and does not affect the model's capability. This, eventually, results in better classification results. 

The parameters of MENET are fine-tuned using a separated set of examples, namely the development  set. During training, the classification accuracy of the model on the development set is continuously monitored. If this metric does not improve for a pre-defined amount of consecutive steps $T_{val}$, the training process is stopped. By using the same mechanism, the learning rate is also annealed when the training proceeds.

\subsubsection{Testing MENET}
To predict the location of users from the test set, we use the trained MENET model to classify these users into pre-defined classes (regions).
The exact geocoordinates of a user is given by the centroid of the respective region. 
The performance of the MENET model is measured by either the accuracy in case of regional classification or distance error metrics (see Section~\ref{performance_criteria}) in case of geographical coordinates prediction.

\subsection{Multiview Features}


Figure~\ref{fig:model1} shows the capability of MENET in exploiting data from multiple sources.
In the context of Twitter user geolocation, we realize MENET by leveraging features from  textual information (\textit{Term Frequency - Inverse Document Frequency}~\cite{MassiveDataMining}, \textit{doc2vec}~\cite{mikolov13sccd}), user interaction network (\textit{node2vec}~\cite{grover2016node2vec}) and metadata (\textit{timestamp}). These features are all extracted from tweets provided they are available. The rest of this section will describe these features and how  they are computed.

\subsubsection{The Term Frequency - Inverse Document Frequency \mbox{Feature}}
The Term Frequency - Inverse Document Frequency (TF-IDF) is a statistical measure used to evaluate how important a term is to a document in a collection or corpus. 
The importance increases proportionally to the number of times the term appears in the document but is offset by the frequency of the term in the corpus. 
TF-IDF is composed of two components, presented next.

\noindent \textit{Term Frequency (TF)}: It measures how often a term occurs in a document. 
The simplest choice is the raw frequency of the term in a document
\begin{equation}\label{tf}
\text{TF}(t,d) = f_{t,d},
\end{equation}
where $f_{t,d}$ is the frequency of term $t$ in the document $d$.

\noindent \textit{Inverse Document Frequency (IDF)}: It measures the informative quantity a term brings across documents. 
Concretely, a common term across multiple documents will be given a low weight while a rare term  will have a higher weight.
The IDF is defined as
\begin{equation}\label{idf}
\text{IDF}(t,D) = \log(\frac{1 + |D|}{1 + \bigl|{d \in D|t \in d}\bigr|}) + 1,
\end{equation}
with $D$ denoting the whole set of documents. 

\noindent Then, the TF-IDF is defined by:
\begin{equation}\label{tfidf}
\text{TF-IDF}(t,d,D) = \text{TF}(t,d) \ \cdot \ \text{IDF}(t,D)
\end{equation}
The output from~\eqref{tfidf} is normalized with the $\ell_2$ norm to have unit length.
In fact, there are many variants for the definition of TF-IDF, and selecting one form depends on the  specific situation. 
We use the formulations~\eqref{tf} and~\eqref{idf} following the existing implementation in the well-established library \textit{scikit-learn}\footnote{http://scikit-learn.org/stable/}~\cite{scikit-learn}. 

\subsubsection{The Context Feature}
The context feature is a mapping from a variable length block of text (e.g. sentence, paragraph, or  entire document) 
to a fixed-length continuous valued vector. 
It provides a numerical representation capturing the context of the document. 
Originally proposed in~\cite{quocle2014},
the context feature is also referred to as \textit{doc2vec} or \textit{Distributed Representation of Sentences},
and it is an extension of the broadly used \textit{word2vec} model~\cite{mikolov13sccd}.

The intuition of \textit{doc2vec} is that a certain context is more likely to produce some sets of words than other contexts. \textit{Doc2vec} trains an embedding capable of expressing the relation between the context and the corresponding words. To achieve this goal, it employs a simple neural network architecture consisting of one hidden layer without an activation function. 
A text window samples some nearby words in a document; 
some of these words are used as inputs to the network and some as outputs. 
Moreover, an additional input for the document is added to the network bringing the document's context. 
The training process is totally unsupervised. After training, the fixed representaion of the document  input will  capture the context of the whole document. 
Two architectures were proposed in~\cite{quocle2014} to learn a document's representation, namely, Distributed Bag of Words (PV-DBOW) and Distributed Memory (PV-DM) versions of Paragraph Vector. 
Athough PV-DBOW is a simpler architecture, it has been claimed that PV-DBOW performs robustly if trained on large datasets~\cite{lau2016}. 
Therefore, we select PV-DBOW model to extract the context feature.

In this paper, we train PV-DBOW models using the \textit{tweets} from the training sets. 
Later, we extract the context feature vectors for the training, 
the development and the test sets. 
Our implementation is based on \textit{gensim}\footnote{https://radimrehurek.com/gensim/}~\cite{rehurek_lrec}.

\subsubsection{The Node2vec Feature}\label{node2vec_feature}
Node2vec is a method  proposed in \cite{grover2016node2vec} 
to learn continuous feature representations (embeddings) for nodes in graphs. 
The low-dimensional feature vector represents the network neighborhoods of a node. 
Let $V$ be the set of nodes of a graph.
\textit{Node2vec} learns a mapping function $f: V \rightarrow \mathbb{R}^d$
that captures the connectivity patterns observed in the graph.
Here, $d$ is a parameter specifying the dimensionality of the feature representation,
and $f$ is a matrix of size $| V | \times d$.
For every source node $v$, a set of neighborhood nodes $N_S(v)\subset V$ is generated through a neighborhood sampling strategy $S$.
Then,  $f$ is obtained by maximizing the log-probability  of observing the neighborhood $N_S(v)$, that is,
\begin{equation} \label{eq:node2vec_target}
\max_f \quad \sum_{v \in V} \log{Pr(N_S(v)|f(v))}.
\end{equation}
%

\textit{Node2vec} employs a sampling method referred to as \textit{biased Random Walk}~\cite{grover2016node2vec}, which samples nodes belonging to the neighborhood of node $v$, 
according to discrete transition probabilities between the current node $v$ and the next node $w$.
These probabilities depend on the distance between the previous node  $u$  and the next node $w$.
Denote by $d_{uw}$ the distance in terms of number of edges from node $u$ to node $w$, 
if the next node coincides with the previous node, then $d_{uw}=0$. 
If the next node has a direct connection to the previous node, then $d_{uw}=1$, 
and if the next node is not connected to the previous node, then $d_{uw}=2$. 
The transition probabilities are defined as follows~\cite{grover2016node2vec}:
\begin{equation}\label{node2vec_transition}
Pr_{vw} = \begin{cases}
\frac{1}{p}, \quad \text{if} \quad d_{uw} = 0, \\
1, \quad \text{if} \quad d_{uw} = 1, \\
\frac{1}{q}, \quad \text{if} \quad d_{uw} = 2,
\end{cases}
\end{equation}
where the parameters $p$ and $q$ are small positive numbers. 
The random walk sampling runs on nodes to obtain a list of walks. 
Later, the node's embeddings are found from the set of walks using the stochastic gradient descent procedure.

\begin{figure}[t]
\centering
\includegraphics[scale=0.45]{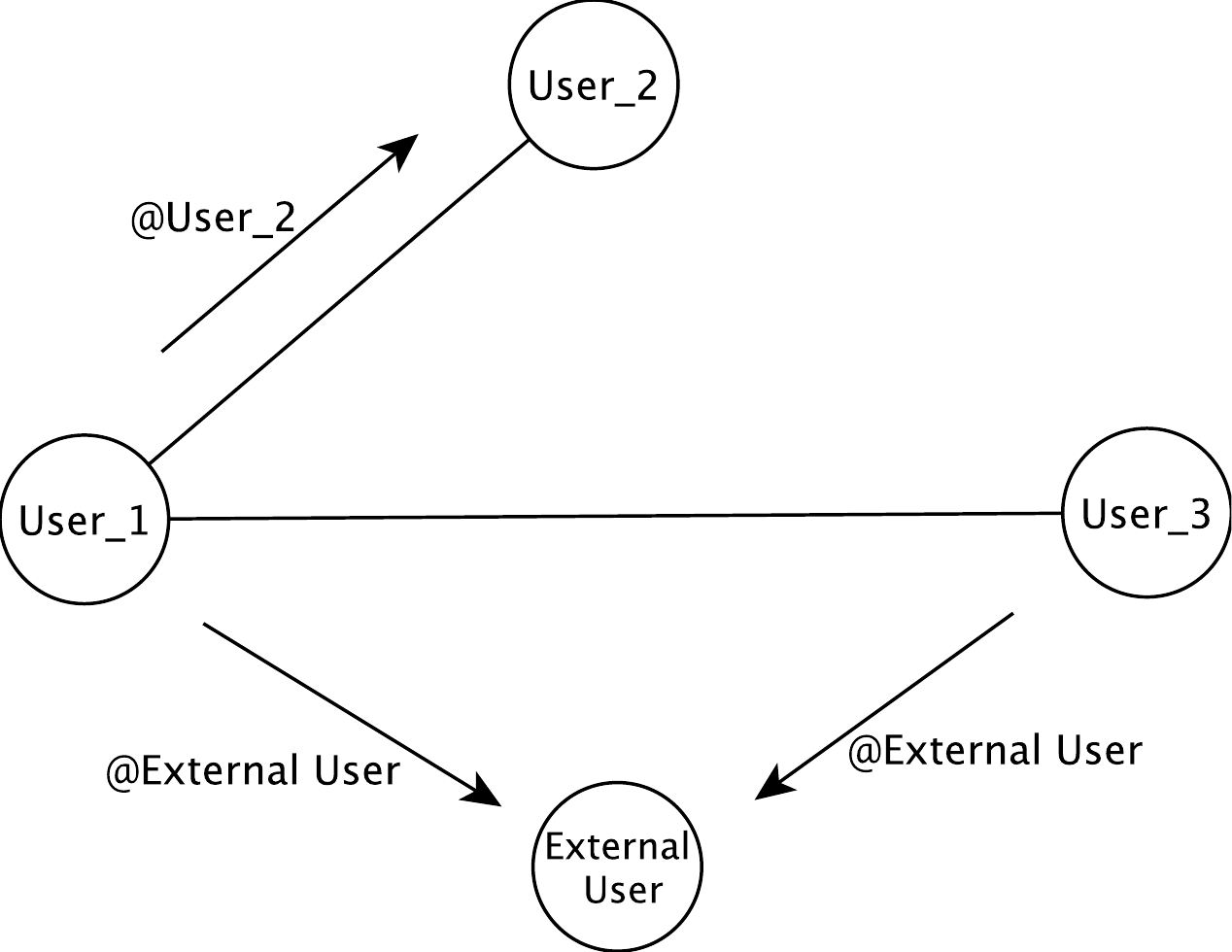}
\centering
\caption{Twitter user graph via mentioning. User\_1 mentions User\_2, thus, we create an edge between them. User\_1 and User\_3 both mention External User, therefore, we make a connection between them.}
\label{fig:graph}
\end{figure}

 
In the context of Twitter user geolocation, each node corresponds to a user, while an edge is the connection between two users. 
We can define these connections by several criteria depending on the availability of data. 
For example, we may consider that two users are connected when actions such as following, mentioning or retweeting are detected. 
In this paper, the content of tweet messages is used to build graph connections. 
Similar to~\cite{rahimi2017},~\cite{rahimi2015}, we 
construct an undirected user graph by employing mention connections. 
First, we create a unique set $V$ with all the users of interest. 
If a user mentions directly another user and both of them belong to $V$, we create an edge reflecting this interaction. 
The edge is assigned a weight equal to the number of mentions. 
To avoid sparsity of the connections, if two users of interest mention a third user, who does not belong to $V$, 
we create an edge between these two users. 
Again, the weight of this edge is the sum of mentions between the third user and the two others. 
Furthermore, we define a list of so-called \textit{celebrities} consisting of users that have a number of unique connections exceeding a threshold $C$. 
We remove all connections to these \textit{celebrities} since the \textit{celebrities} are often mentioned by plenty of people all over the world. Mentioning a \textit{celebrity}, therefore, might not be a good indication of geographical relation.
The graph building procedure is depicted in Fig.~\ref{fig:graph}.

A shortcoming of this method is that it can only produce an embedding for a node   
if that node has at least one connection to another node. 
Nodes without an edge can not be represented. 
Therefore, for an isolated node, we consider an all-zero vector as its embedding. 
Moreover, whenever a new node joins the graph, 
the algorithm needs to run again to learn feature vectors for all the nodes of the graph, 
making our method inherently transductive. 
There are some existing efforts addressing this problem. 
In~\cite{yangcs16}, the authors consider a node's embedding as a function of its \emph{natural feature};
in this case the embedding could be a function of either the TF-IDF or \textit{doc2vec} feature. 
A similar approach presented in \cite{hamiltonyl17} 
generates a node's embedding by sampling and aggregating features from the node's local neighborhood. 
These inductive approaches will be considered in our future work.

\subsubsection{The Timestamp Feature}
In many commonly used Twitter databases like GeoText~\cite{eisenstein2010} and UTGeo2011~\cite{roller2012}, 
the posting time of all tweets is available in UTC value (Coordinated Universal Time). 
This allows us to leverage another view of the data. 
In~\cite{dredze2016}, it was shown that there exists a correlation between time and place in a Twitter stream of data. 
In fact, it is less likely that people tweet late at night than at any other time, which implies a drift in longitude. 
Therefore, the \textit{timestamp} could be an indication for a time zone. 
We obtain the \textit{timestamp} feature for a given user as follows.
First, we extract the timestamps from all the tweets of that user
and convert them to the standard format to extract the hour value.
Then, a $24$-dimensional vector is created corresponding to $24$ hours in a day;
the $i$-th element of this vector equals the number of messages posted by the user at the  $i$-th hour. 
This feature is $\ell_2$ normalized to a unit vector before feeding it to our neural network model.

\subsection{Improvements with S2 adaptive grid}
\label{subsection:s2_adaptive_grid}
When addressing the prediction of users' location as a classification problem,
the geographical coordinate assigned to a user with unknown location 
equals the centroid of the class, which has been predicted for the user. A straightforward way to form the classes is taking administrative boundaries such as states, regions or countries.
Such an approach brings large distance errors if the respective areas are large. 
Intuitively, the prediction accuracy could be improved  
if we increase the granularity level by defining classes that correspond to smaller areas. 
The tiling should also consider the distribution of users; 
very imbalanced custom classes should be avoided,
otherwise, the training process will not be efficient. 
Therefore, finding an appropriate way to subdivide users 
into custom small geographical areas is critical. 

An early work of Roller \textit{et al.}~\cite{roller2012} 
has built an adaptive grid using a \textit{k}-d tree to partition data into custom classes. 
This partitioning, though considers the distribution of users, does not necessarily produce uniform cells at the same level. 
Here, we split the Twitter users in the training set into small areas called S2 cells, 
using Google's S2 geometry library. 
This library is a powerful tool for partitioning the earth's surface. 
Considering the earth as a sphere, the library hierarchically subdivides the sphere's surface by projecting it on an enclosing cube. 
On each surface of the cube, a hierarchical partition is made using a spatial data structure named \textit{quad-tree}. 
Each node on the tree represents an S2 cell, which corresponds to an area on the earth's surface. 
The quad-tree used in the Google S2 geometry library 
has a depth of $30$; the root cell is assigned the lowest level of zero and the leaf cells are assigned the highest level of $30$. 
The library outputs mostly uniform cells at the same level.
For instance, the minimum area of level-12 cells is $3.31$~km\textsuperscript{2} 
and the maximum area of these cells is $6.38$~km\textsuperscript{2}. 

\begin{figure*}[t]
\centering
\begin{tabular}{c c}
\includegraphics[width=80mm]{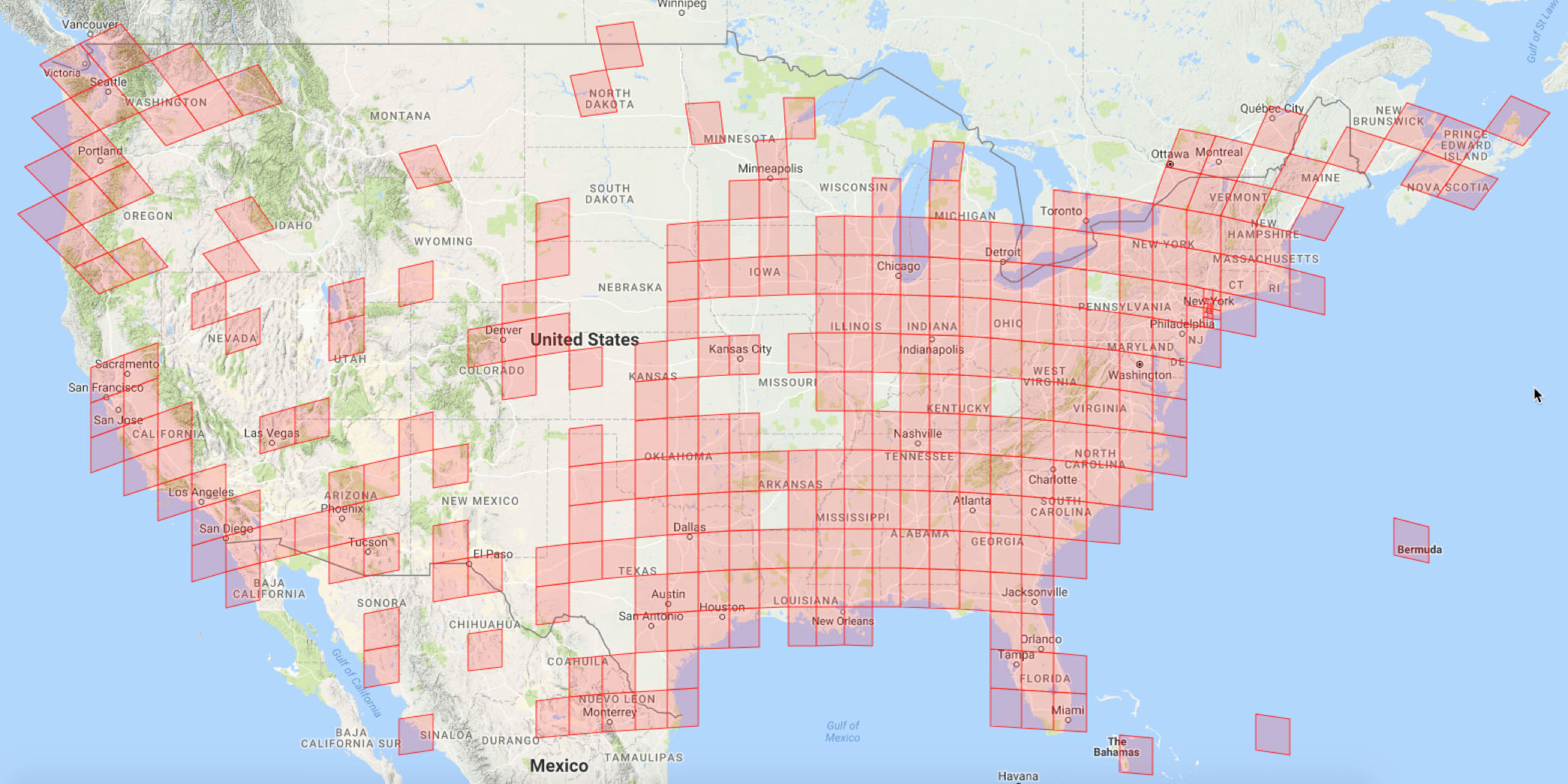} & \includegraphics[width=80mm]{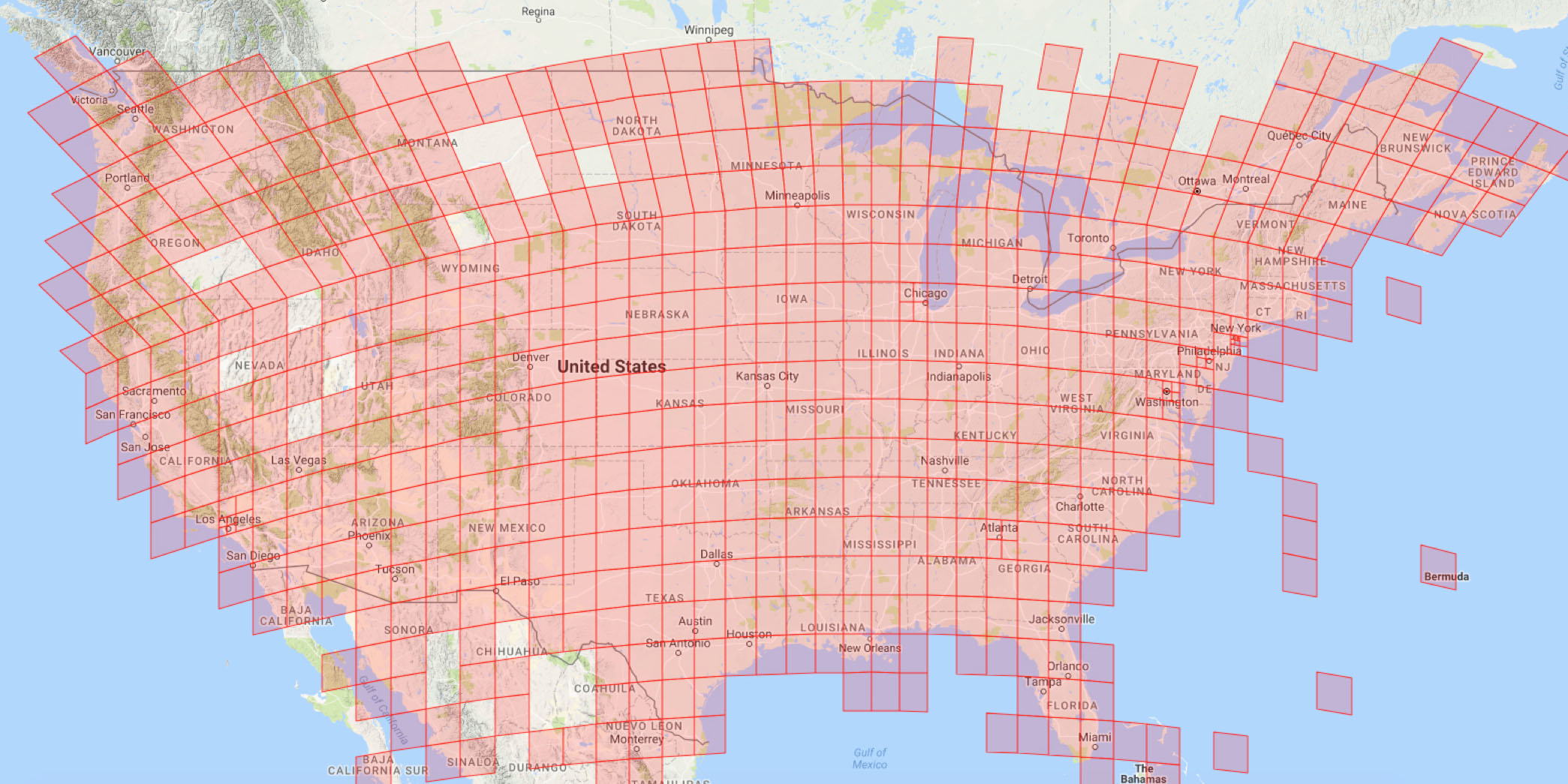} \\
(a) & (b) 
\end{tabular}
\caption{Partitioning Twitter users with S2 cells for (a) GeoText~\cite{eisenstein2010} and  (b) UTGeo2011~\cite{roller2012}. $L_{min}$ is set to $6$ for both datasets while $T_{max}$ is set to $500$ and $10.000$ for GeoText and UTGeo2011, respectively. Highly densed cities, like New York, are split in small cells while most of other regions reach $L_{min}$ because of the small amount of users. The tiling does not cover the whole US area because there are regions without tweets.}
\label{fig:tiling}
\end{figure*}

\begin{figure*}
\centering
\includegraphics[width=0.9\linewidth]{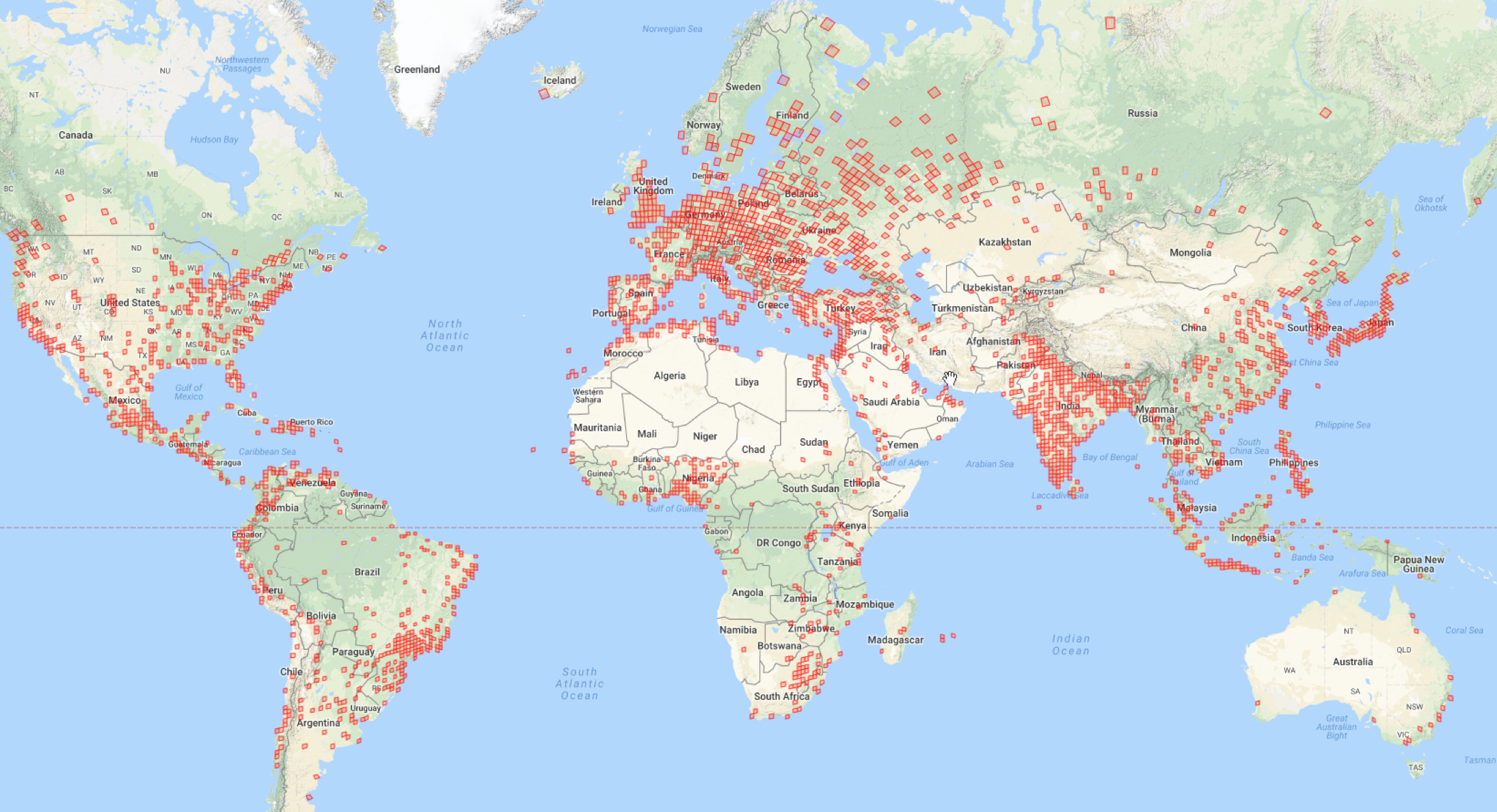}
\caption{Partitioning Twitter users for the TwitterWorld dataset~\cite{bohan_twitter_world_2012} with S2 cells created with $L_{min} = 7$ and $T_{max} = 50.000$.}
\label{fig:world_tiling}
\end{figure*}

In this work, we build an S2 adaptive grid, aiming at a balanced tiling,
meaning that the defined cells (geographical areas)  contain a similar number of users.
For this reason, we specify a threshold $T_{\max}$, as the maximum allowed number of users per cell.
We build the adaptive grid from bottom to top. First, we identify the leaves corresponding to given geocoordinates.
As long as the total number of users in children nodes (cells) is smaller than $T_{\max}$,
we merge these nodes together; the children nodes' users are assigned to the parent cell, i.e., 
 a larger geographical area.
We climb the tree gradually repeating this process.
If we reach a specific level, $L_{\min}$, we stop the climb 
in order to avoid defining cells that correspond to large geographical areas;
otherwise,  the prediction error  would increase.
%
%
Figures~\ref{fig:tiling} and~\ref{fig:world_tiling} show the subdivision of users in S2 cells for the considered datasets.

\section{Experiments}\label{experiments}
\subsection{Datasets}
In our experiments, we employ the following three datasets, which contain tweets coming from the United States (GeoText~\cite{eisenstein2010}, UTGeo2011~\cite{roller2012}) and all over the world (TwitterWorld~\cite{bohan_twitter_world_2012}).\\

\noindent \textit{GeoText:} This is a small dataset containing more than $370.000$ tweets posted by  $9475$ unique users from $48$ contiguous states and Washington D.C.~during the first week of March, 2010. 
Tweets were filtered carefully before being put into the dataset to make sure that only relevant tweets are kept. 
In this dataset, the geospatial coordinates of the first message of users were used as their primary location. 
This was done originally by the author in~\cite{eisenstein2010} and followed by other authors~\cite{roller2012,rahimi2015}. 
The dataset was already split into the training, development and testing sets with $7580$, $1895$ and $1895$ users, respectively.  
For the downstream tasks, tweets from a user are concatenated making a \textit{tweet document}. \\

\noindent \textit{UTGeo2011:} This is a larger dataset which was created by the authors of~\cite{roller2012}. The dataset is also referred to as \textit{TwitterUS} in many Twitter user geolocation publications~\cite{rahimi2015,rahimi2017,melo2015geocoding}.  
The dataset contains approximately $38$ million tweets sent by $449.694$ users from the US. 
In contrast to GeoText, this dataset is noisier, namely 
 many tweets have no location information. 
To treat it similarly to GeoText, all the tweets from a specific user are concatenated into a single document;  
a primary location is defined as the earliest valid coordinate of the tweets. 
Ten thousand users are selected randomly to make the development set, 
and the same amount is reserved for the evaluation set. 
The remaining users form the training set. \\

\noindent \textit{TwitterWorld:} This is the dataset created by the authors of~\cite{bohan_twitter_world_2012}. The dataset contains $12$ million tweets sent by $1.39$ million users from different countries in the world, of which ten thousand users are kept for each the development set and the testing set. Moreover, only tweets that are in English and close to a city are retained. The primary location of a user in this dataset is assigned the centre of the city where most of his tweets were sent. Different from  GeoText and UTGeo2011, this dataset provides purely textual information; the timestamps of messages are not available.

\vspace{0.2cm}
The location of a user is indicated by a pair of real numbers, namely, latitude and longitude. 
However, classification models need discrete labels. For the datasets collected from the US, we follow~\cite{eisenstein2010,roller2012} to employ administrative boundaries to create the class labels. By doing so, we can consider the tasks of regional and state classification as in~\cite{eisenstein2010,LiuPaper,ICWSM1510561}.
We rely on the Ray Casting algorithm of~\cite{shimrat1962algorithm} to decide if a location is inside a region or state's  boundary. 
For the region and state boundaries, we use information from Census Divisions\footnote{https://www2.census.gov/geo/pdfs/maps-data/maps/reference/us\_regdiv.pdf}.
Also, we have employed the Google S2 library, \textit{k}-means and \textit{k}-d tree clusterings to partition all the geospatial datasets, making other sets of labels. This supports the task of predicting the geocoordinates. More details on the settings of the partitioning schemes and their impacts will follow in the next section.

\subsection{Performance Criteria and Experiment Design}
\label{performance_criteria}
The proposed model for geolocation of Twitter users addresses the following tasks:
(i) four-way classification of US regions including Northeast, Midwest, West and South, 
(ii) fifty-way classification to  predict the states of users,
and (iii) estimation of the real-valued coordinates of users, i.e., latitude and longitude. 
For the region and state classification tasks, 
we compare the performance of our model with existing methods,
by calculating the percentage of correctly classified users, which is the accuracy. 
Considering the estimation of the user coordinates, 
we measure the distance between the predicted and the actual geocoordinates 
and calculate the mean and the median values over the testing dataset.
The distance between the predicted and the ground truth coordinates is computed using the Haversine formula~\cite{haversine1984}.
Another common way to measure the success of coordinate estimation
is to calculate the percentage of estimations with accuracy better than $161$ km;
this metric, known as \textit{@$161$}\footnote{$161$~km $\sim$ $100$ mile}, has been used in many works~\cite{roller2012,wing2011geodesic,wing2014hierarchical,rahimi2015,rahimi2017,LiuPaper}. 
It is worth noting that for the classification accuracy and the accuracy @$161$ metrics, the higher values indicate a good prediction. Conversely, achieving lower values for the mean and median distance errors is desired.

Concerning the first two classification tasks, we conduct experiments on the US Twitter datasets, namely GeoText and UTGeo2011. For predicting Twitter users' geocoordinates, experiments are performed on the three datasets.  Furthermore, it should be noted that the experiments for geographical coordinate prediction use different sets of labels created by S2, $k$-d tree and $k$-means partitioning. Also, the administrative boundaries used in task (ii) are exploited for exact geocoordinate estimation.

\subsection{Data Pre-processing and Normalization}
Before computing \textit{node2vec} and \textit{TF-IDF} features, a simple pre-processing phase is required. 
First, we tokenize the tweets and remove stop words using \textit{nltk}\footnote{http://www.nltk.org/}~\cite{bird2009}, 
a dedicated library for natural language processing. 
Then, we replace URLs and punctuation by special characters, 
which results in reducing the size of the vocabulary without affecting the semantics of tweets. 
Again, \textit{nltk} is used for stemming in the last stage of pre-processing.

Normalization is a common step to pre-process data before applying machine learning algorithms. 
Data can be normalized by removing the mean and dividing by the standard deviation. 
Alternatively, samples can be scaled into a small range of $[0,1]$ or $[-1,1]$. 
The less common way is to scale the samples so that their module is equal to 1, also known as $\ell_2$ normalisation.
In our case, the TF-IDF, node embedding and context features are already scaled to the range [0,1]. We apply $\ell_2$ normalization for the \textit{timestamp} feature only.

\subsection{Parameter Settings}
Our framework considers four different features and each feature requires some parameters for extraction. Extracting \textit{TF-IDF} using \textit{scikit-learn} requires a minimum term frequency across documents \textit{min\_df}. For the GeoText dataset, we choose \textit{min\_df}=$40$. For the  UTGeo2011 and TwitterWorld datasets, because of the sheer volume of data, we set \textit{min\_df}=$500$ and \textit{min\_df}=$400$, respectively. Concerning \textit{doc2vec}, we select an embedding size equal to $300$. The size of the sampling window is set to $10$.

We have built the Twitter users' graphs for the three datasets using mentions extracted from tweet messages only as discussed in Section~\ref{node2vec_feature}. Following~\cite{rahimi2015}, we set the celebrity connection thresholds $C$ to $5$, $15$ and $5$ for GeoText, UTGeo2011 and TwitterWorld, respectively. Table~\ref{table:graph_statistics} shows graph statistics for all three  datasets.
We use the code provided by the authors of~\cite{grover2016node2vec} to obtain the  \textit{node2vec} feature. We choose an embedding size equal to $300$. When training the embeddings, we select the weighted graph option, which takes into account the weights of edges. Other parameters are set to default values, namely the walk length $l=80$, transition parameters $p=1$, $q=1$. The sampling window size is set to $5$.

\begin{table}[t]
\centering
\caption{Statistics of Twitter users' graphs.}
\label{table:graph_statistics}
\begin{tabular}{| c | c | c | c |}
\hline
 & GeoText & UTGeo2011 & TwitterWorld \\
 \hline
 \hline
 Node count & 9475 & 449.508 & 1.386.766 \\
 \hline
 Edge count & 55.640 & 5.297.215 & 1.076.462 \\
\hline
\end{tabular}
\end{table}

%

\begin{table}[h]
\centering\caption{Hyperparameter setting for MENET with regard to region/state classification and  geocoordinates prediction. $n_{h_{11}}, n_{h_{21}}, n_{h_{31}}, n_{h_{41}}$ are the numbers of neurons in the hidden layers $h_{11}, h_{21}, h_{31}, h_{41}$ for the \textit{TF-IDF}, \textit{doc2vec}, \textit{node2vec}, and \textit{timestamp} features, respectively.}
\label{table:menet_parameters}
\begin{tabular}{| c | c | c |}
\hline
& Region/State classification & Coordinates Prediction \\
\hline
\multirow{ 3}{*}{Datasets} & GeoText & GeoText \\
& UTGeo2011 & UTGeo2011 \\
& & TwitterWorld \\
\hline
\hline
$n_{h_{11}}$ & 150 & 100 \\ 
\hline
$n_{h_{21}}$ & 150 & 300 \\ 
\hline
$n_{h_{31}}$ & 30 & 300 \\ 
\hline
$n_{h_{41}}$ & 30 & 100 \\ 
\hline
\end{tabular}
\end{table}

Choosing the right hyperparameters for neural networks, which are the number and size of hidden layers, is always a challenge. In our experiments, these parameters are set empirically. 
We set the number of hidden layers on each individual branch to 1, namely we use hidden layers $h_{11}$,  $h_{21}$,  $h_{31}$ and $h_{41}$ for features TF-IDF, \textit{node2vec}, \textit{doc2vec}, and \textit{timestamp}, respectively. Also, we connect the combination layer with the softmax layer directly without adding any layer in between.
All hyperparameters can be found in Table~\ref{table:menet_parameters}. We use a small value for the learning rate $\alpha=0,0001$ and regularize the weights right before the output layer only. The regularization parameter $\lambda$ is set to $0,1$. The training procedure is performed using stochastic gradient descent with the optimization algorithm ADAM~\cite{kingma2014} as the updating rule. The consecutively non-improving performance threshold $T_{val}$ is set to $10$ for GeoText and $6$ for both UTGeo2011 and TwitterWorld datasets.

Creating S2 grids requires setting the minimum cell level $L_{min}$ and maximum number of users per cell $T_{max}$. We have experimented with different settings and reported the best result in Table~\ref{table:coordinates_prediction} with $L_{min} = 6$,  $T_{max} = 500$ for GeoText, $L_{min} = 6$, $T_{max} = 10.000$ for UTGeo2011 and $L_{min} = 7$, $T_{max} = 50.000$ for TwitterWorld.

\subsection{Results}
After experimenting with different parameters, 
normalization techniques and feature combination strategies,
we report here the best obtained results.
Table \ref{table:classification_result} presents results for regional and state geolocation for GeoText and UTGeo2011,
while for the prediction of user geographical coordinates, results are presented in Table \ref{table:coordinates_prediction}. 

Concerning the classification tasks, our model significantly outperforms all previous works. 
Successful regional classification is achieved for  $76\%$ of users, 
while for state classification the result is $64.4\%$. 
By leveraging the classification strength of multiple features, 
the improvement in regional accuracy is $9\%$ compared to the work in~\cite{ICWSM1510561}. 
Concerning the accuracy in state classification, 
we achieve a greater improvement that rises to $23\%$ 
compared to the state of the art presented in~\cite{ICWSM1510561}.
\begin{table}[t]
\centering
\caption{Regional and state classification results on GeoText and UTGeo2011. N/A stands for not available.}
\label{table:classification_result}
 \begin{tabular}{|c | c c | c c|} 
 \hline
  & \multicolumn{2}{|c|}{GeoText} & \multicolumn{2}{|c|}{UTGeo2011} \\
 \hline
  & Region  & State & Region  & State \\
  & (\%) & (\%) & (\%) & (\%) \\
\hline
\hline
Eisenstein \textit{et al.}~\cite{eisenstein2010} & 58 & 27 & N/A & N/A \\
\hline
Cha \textit{et al.}~\cite{ICWSM1510561} & 67 & 41 & N/A & N/A \\
\hline
Liu \& Inkpen~\cite{LiuPaper} & 61.1 & 34.8 & N/A & N/A \\
\hline
MENET & \textbf{76} & \textbf{64.4} & 83.7 & 69 \\ 
\hline
\end{tabular}
\end{table}

\begin{table*}[t!]
\centering
\caption{Performance comparison on geographical coordinates prediction. N/A stands for not available.}
\label{table:coordinates_prediction}
 \begin{tabular}{| c | c c c | c c c | c c c |}
 \hline
  & \multicolumn{3}{|c|}{GeoText} & \multicolumn{3}{|c|}{UTGeo2011} & \multicolumn{3}{|c|}{TwitterWorld} \\ [0.5ex] 
 & mean & median & @161 & mean & median & @161& mean & median & @161 \\
  & (km) & (km) &(\%) &(km) &(km) &(\%) &(km) &(km) &(\%) \\
\hline
\hline
Eisenstein \textit{et al.}~\cite{eisenstein2010} & 900 & 494 & N/A & N/A & N/A & N/A & N/A & N/A & N/A\\

Wing \textit{et al.}~(2011)~\cite{wing2011geodesic} & 967 & 479 & N/A & N/A & N/A & N/A & N/A & N/A & N/A\\ 

Roller \textit{et al.}~\cite{roller2012}  & 897 &432 & 35.9 & 860 & 463 & 34.6 & N/A & N/A & N/A \\

Wing \& Baldridge~(Uniform)~\cite{wing2014hierarchical} & N/A & N/A & N/A & 703.6 & 170.5 & 49.2 & 1714.6 & 490 & 32.7 \\

Wing \& Baldridge~(KD tree)~\cite{wing2014hierarchical} & N/A & N/A & N/A & 686.6 & 191.4 & 48.0 & 1669.6 & 509.1 & 31.3 \\

Melo \textit{et al.}~\cite{melo2015geocoding} & N/A & N/A & N/A & 702 & 208 & N/A & 1507 & 502 & N/A \\

Liu \& Inkpen~\cite{LiuPaper} & 855.9 & N/A & N/A & 733 & 377 & 24.2 &  N/A & N/A & N/A \\

Cha \textit{et al.}~\cite{ICWSM1510561} & 581 & 425 & N/A & N/A & N/A & N/A & N/A & N/A & N/A \\

Rahimi \textit{et al.}~(2015)~\cite{rahimi2015} & 581 & 57 & 59 & 529 & 78 & 60 & 1403 & 111 & 53 \\

Rahimi \textit{et al.}~(2017)~\cite{rahimi2017} & 578 & 61 & 59 & 515 & 77 & 61 & 1280 & \textbf{104} & 53 \\ 
 \hline
 MENET with state labels & 570 & 58 & 59.1 & 474 & 157 & 50.5 & N/A & N/A & N/A\\ 

 MENET with S2 labels & \textbf{532} & \textbf{32} & \textbf{62.3} & \textbf{433} & \textbf{45} & \textbf{66.2} & \textbf{1044} & 118 & \textbf{53.3} \\ 
 \hline
\end{tabular}
\end{table*}

The estimation of geographical coordinates of Twitter users 
involves experiments with two types of labels, thus two sets of experiments. 
In the first set of experiments, we use classes corresponding to the fifty states of the US.
In the second set of experiments, we employ the S2 classes described in Section \ref{methodology}.
As can be seen in Table~\ref{table:coordinates_prediction},
concerning the results obtained with state labels, 
the mean distance error obtained with MENET is smaller than with other methods. 
Likewise, the median distance error and the @$161$ accuracy are better on GeoText. However, our result with these metrics is worse on UTGeo2011. 
The reason being that the state boundaries ignore the geographical distribution of users. 
The performance of MENET is improved significantly over all criteria with S2 labels,
when the definition of regions takes into account the distribution of users. 
In this case, Table~\ref{table:coordinates_prediction} shows that 
the proposed method outperforms existing methods
in terms of mean, median distance error and @$161$ accuracy on GeoText and UTGeo2011. On  \mbox{TwitterWorld}, the median distance error is reduced more than $200km$ compared to the result in~\cite{rahimi2017} while the result for the other metrics is comparable to the state of the art.
At this point, we would like to underline that
the number of employed classes is critical for the performance of our method.
A larger number of classes results in smaller geographical areas,
which may improve the geocoordinate prediction.
However, training a model with more classes may be more difficult,
thus, the classification may perform worse.

\subsubsection{Granularity Analysis}

\begin{table}
\centering
\caption{Performance of MENET on GeoText with respect to varying the minimum level of S2 cell $L_{min}$. In these experiments, $T_{max}$ is set to $500$ and $L_{min}$ varies from $3$ to $8$.}
\label{table:menet_performance_vary_level}
 \begin{tabular}{| c | c | c | c | c |}
 \hline
 $L_{min}$ & Region count & Mean (km) & Median (km) & @161(\%) \\
\hline
\hline
3 & 71 & 554 & 71 & 58.6  \\
4 & 89 & 546 & 65 & 59.3\\
5 & 148 & 534 & 47 & 60.7\\
6 & 306 & \textbf{532} & 32 & \textbf{62.3} \\
7 & 590 & 574 & \textbf{28} & 62.0\\
8 & 947 & 706 & 46 & 55.3\\
 \hline
\end{tabular}
\end{table}

\begin{table}
\centering
\caption{Performance of MENET on GeoText with respect to varying the maximum number of users per cell. In these experiments, $L_{min}$ is set to $6$ and $T_{max}$ varies from $100$ to $600$.}
\label{table:menet_performance_very_user}
 \begin{tabular}{| c | c | c | c | c |}
 \hline
$T_{max}$ & Region count & Mean (km) & Median (km) & @161(\%) \\
\hline
\hline
100 & 470 & 1257 & 877 & 25.8 \\
200 & 353 & 581 & 33 & 61.5 \\
300 & 333 & 564 & 33 & 62.1 \\
400 & 318 & 559 & 33 & 62.0 \\
500 & 306 & \textbf{532} & \textbf{32} & \textbf{62.3} \\
600 & 300 & 576 & 35 & 61.5 \\
 \hline
\end{tabular}
\end{table}

As explained in Section~\ref{subsection:s2_adaptive_grid}, an S2 adaptive grid is built using two parameters: the minimum S2 cell level $L_{min}$ and the maximum number of users per cell (region, class) $T_{max}$. As an example, the geolocation result with S2 labels presented in Table~\ref{table:coordinates_prediction} for GeoText is associated with the minimum cell level of $6$ and the user threshold of $500$. The number of cells and their area ($m^2$) will vary depending on these parameters. One may wonder if this setting is optimal or not. In this section, we present an analysis of the performance of MENET with regard to different S2 parameter settings. Concretely, we run experiments using the same hyperparameter setting of MENET on GeoText with different S2 label sets. The label sets are created by either varying the minimum S2 cell level $L_{min}$ or the user threshold $T_{max}$. The results of these experiments are shown in Tables~\ref{table:menet_performance_vary_level} and ~\ref{table:menet_performance_very_user}.

We can see a clear trend in the median of the distance error from the experiments with varying  $L_{min}$. When $L_{min}$ increases, meaning more regions are generated, the median of the  distance error decreases monotonically to a very small value (i.e., $28$ km). The reason for this is very intuitive. S2 cells at a higher level have smaller area, and if the classification performance of MENET does not get significantly worse with more classes, a predicted location will be more likely closer to the ground truth location. This also explains the increasing trend in accuracy within  $161$ km. There is no clear trend in the mean of distance error. This could be explained by the sensitivity of the mean with regard to the outliers. Even if the classification accuracy of MENET goes down slightly, it may bring huge distance errors from large area cells. This has a large impact on the mean value. On the other hand, the impact of these outliers is small on the median value.

\begin{figure*}[htp]
\centering

\begin{tabular}{c c c c c}
\includegraphics[width=30mm]{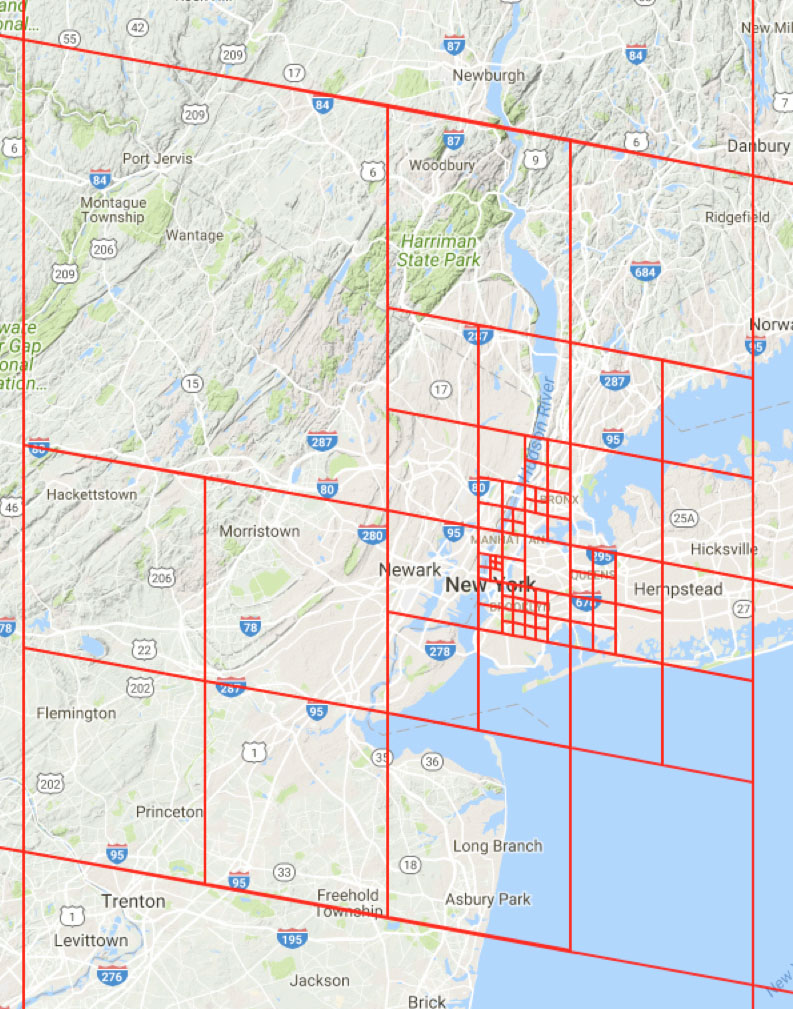} & \includegraphics[width=30mm]{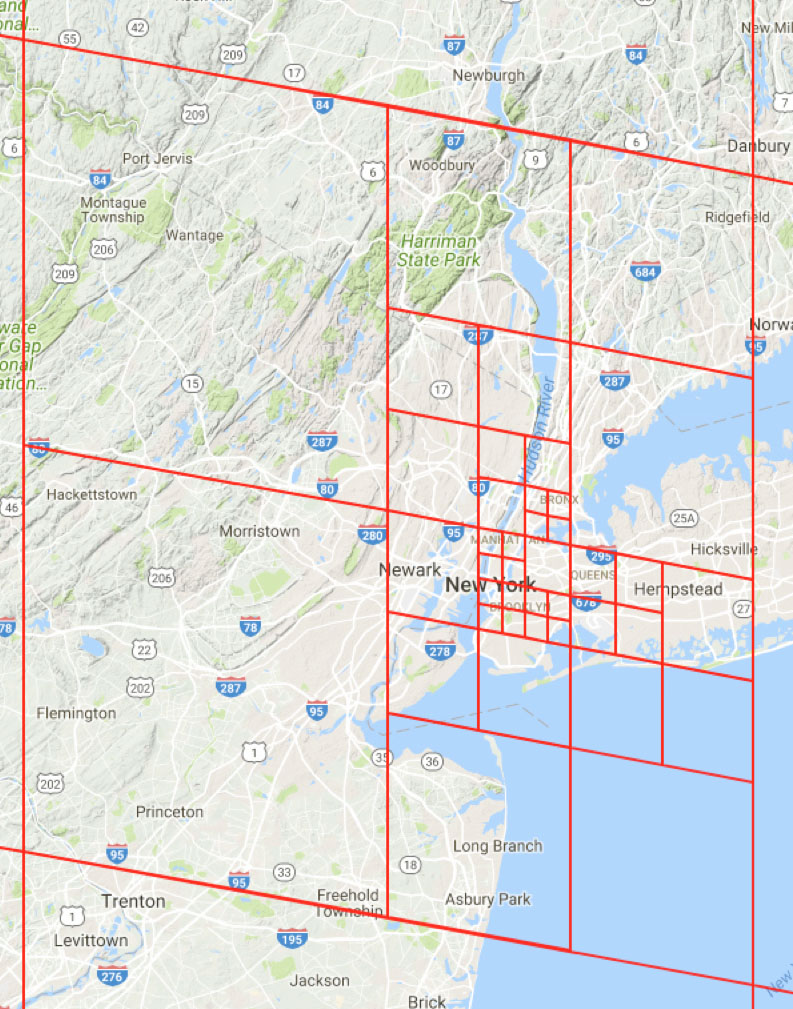} & \includegraphics[width=30mm]{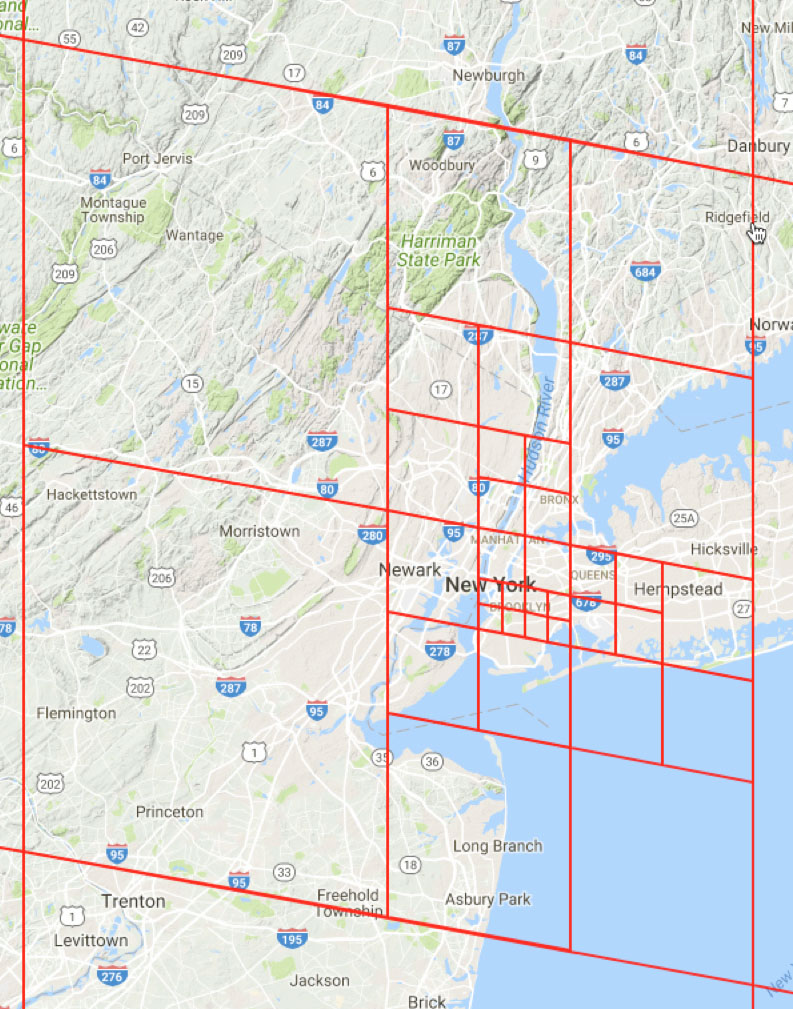} & \includegraphics[width=30mm]{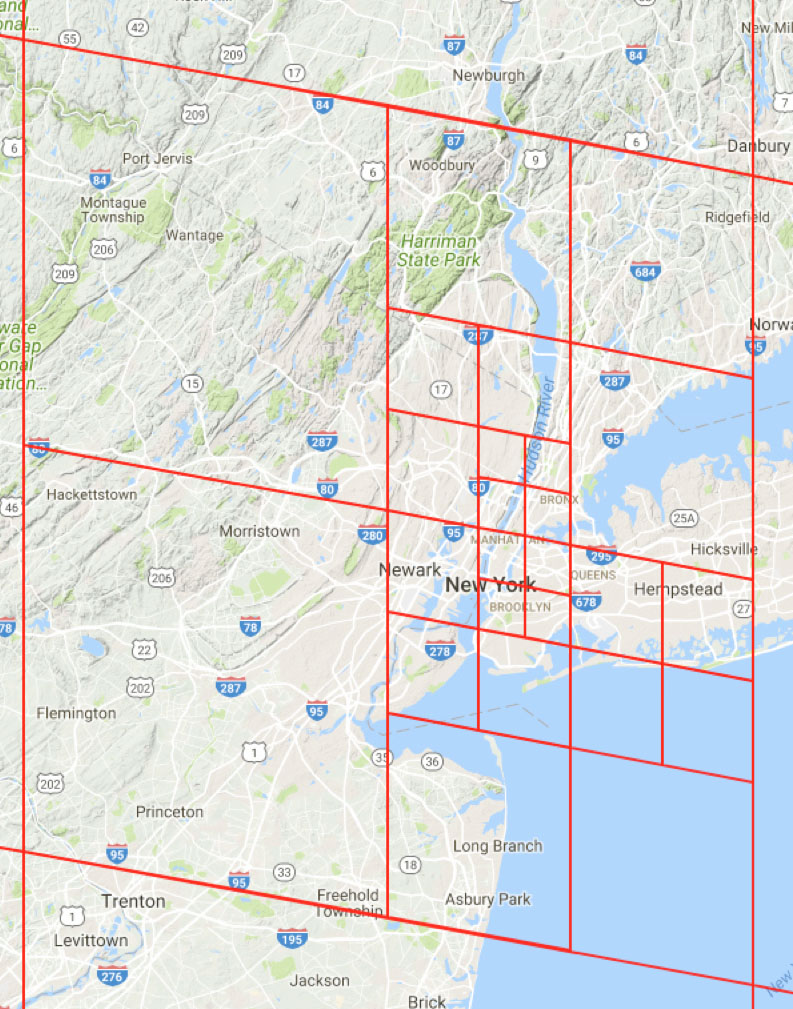} & \includegraphics[width=30mm]{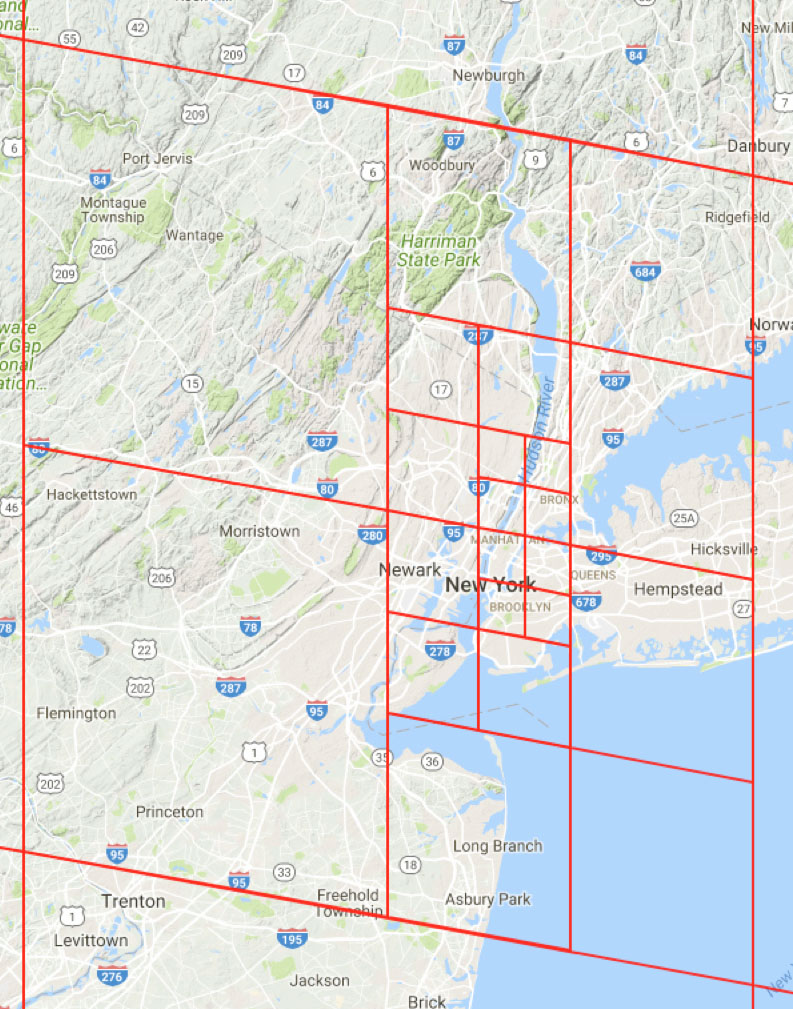} \\
(a) $T_{max} = 100$ & (b) $T_{max} = 200$ & (c) $T_{max} = 300 $ & (d) $T_{max} = 400$ & (e) $T_{max} = 500$\\
\end{tabular}

\caption{Partitioning of the New York region with $L_{min} = 6$. A finer-grained grid can be obtained with smaller user thresholds.}
\label{fig:granularity_newyork}

\end{figure*}

\begin{figure*}[htp]
\centering
\begin{tabular}{c c c c c}
\includegraphics[width=30mm]{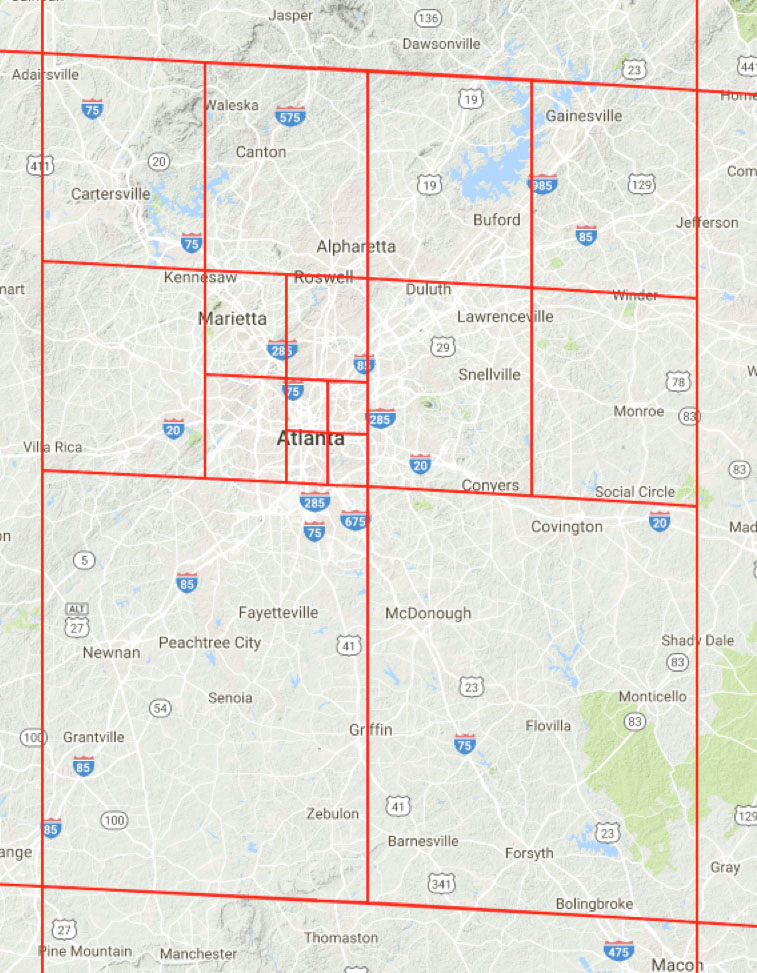} & \includegraphics[width=30mm]{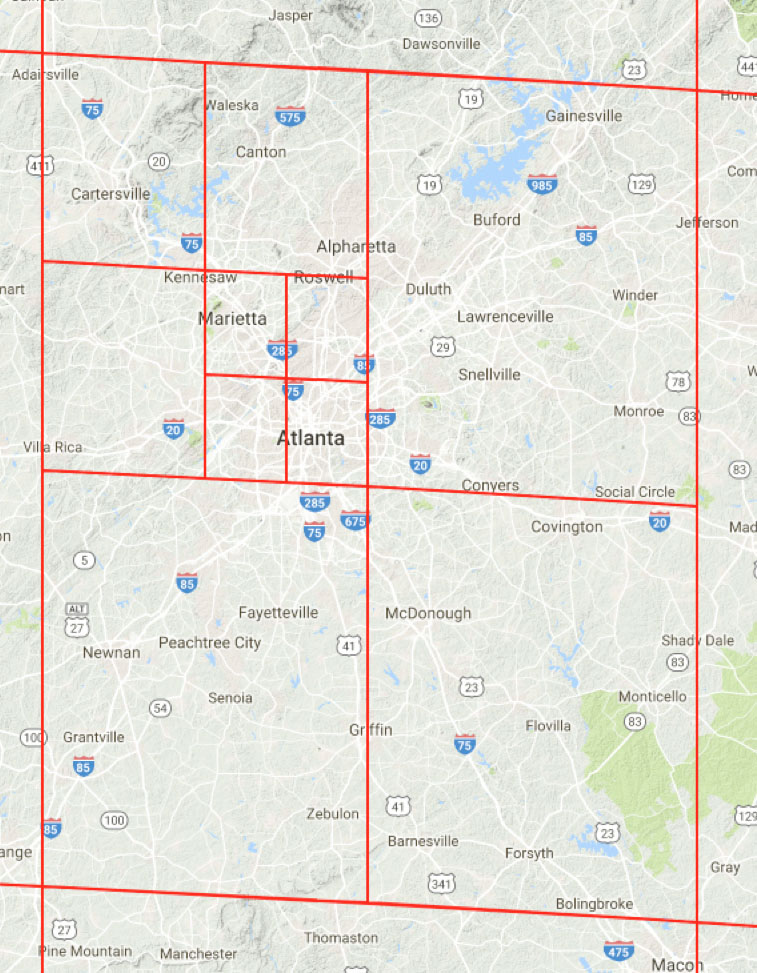} &
\includegraphics[width=30mm]{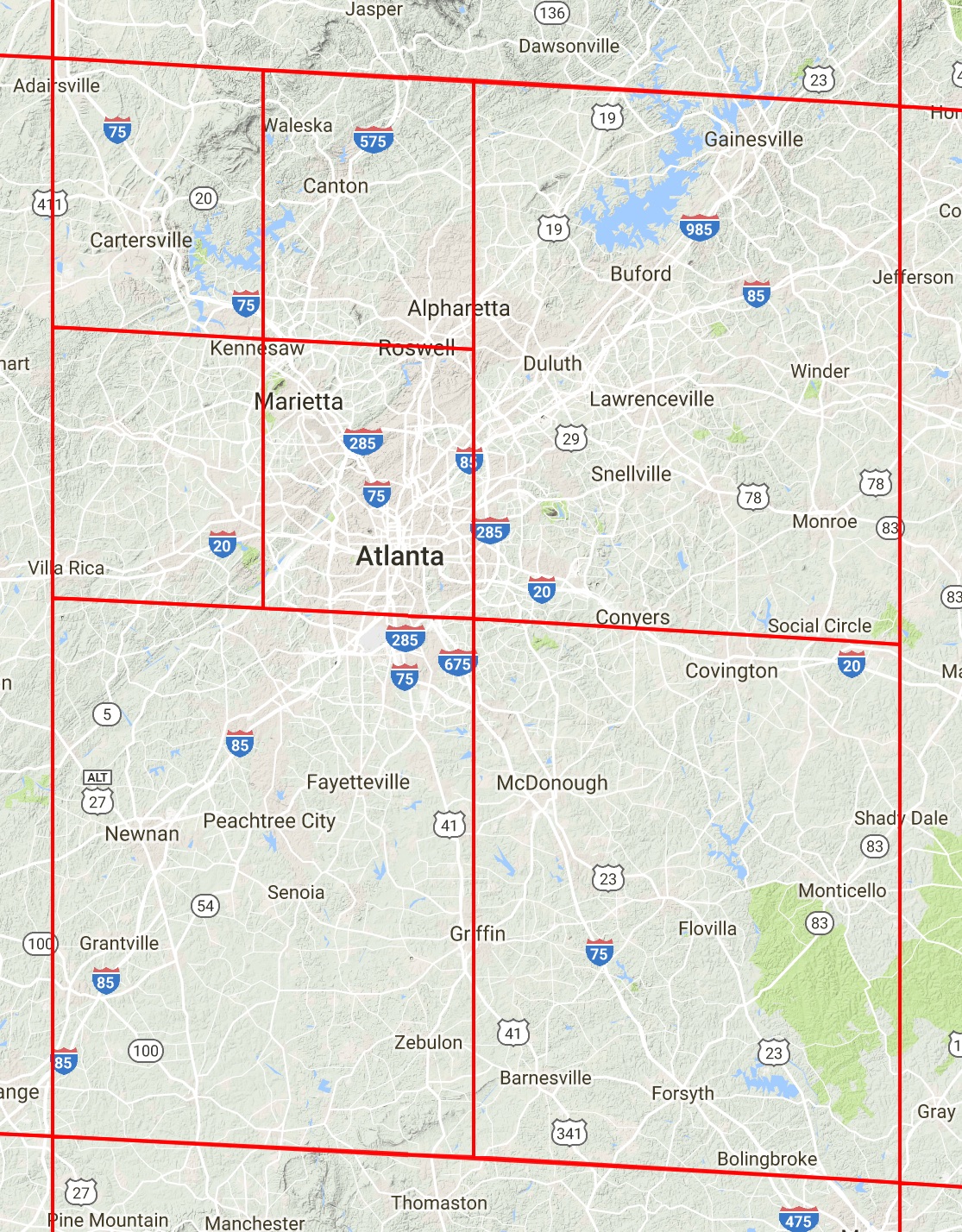} & \includegraphics[width=30mm]{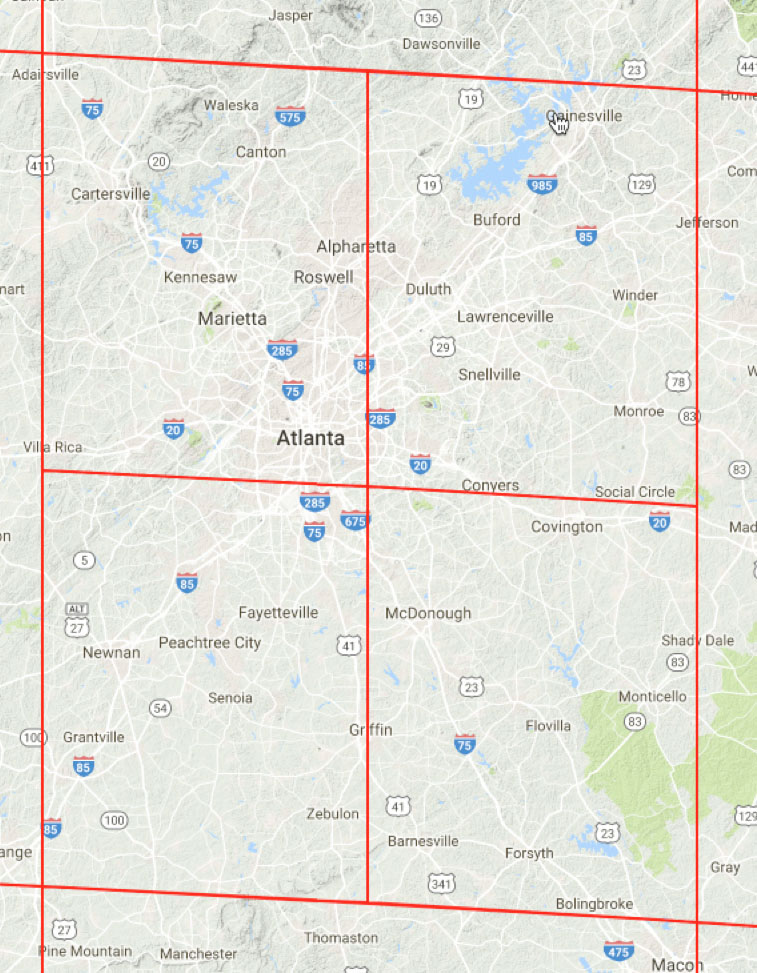} & \includegraphics[width=30mm]{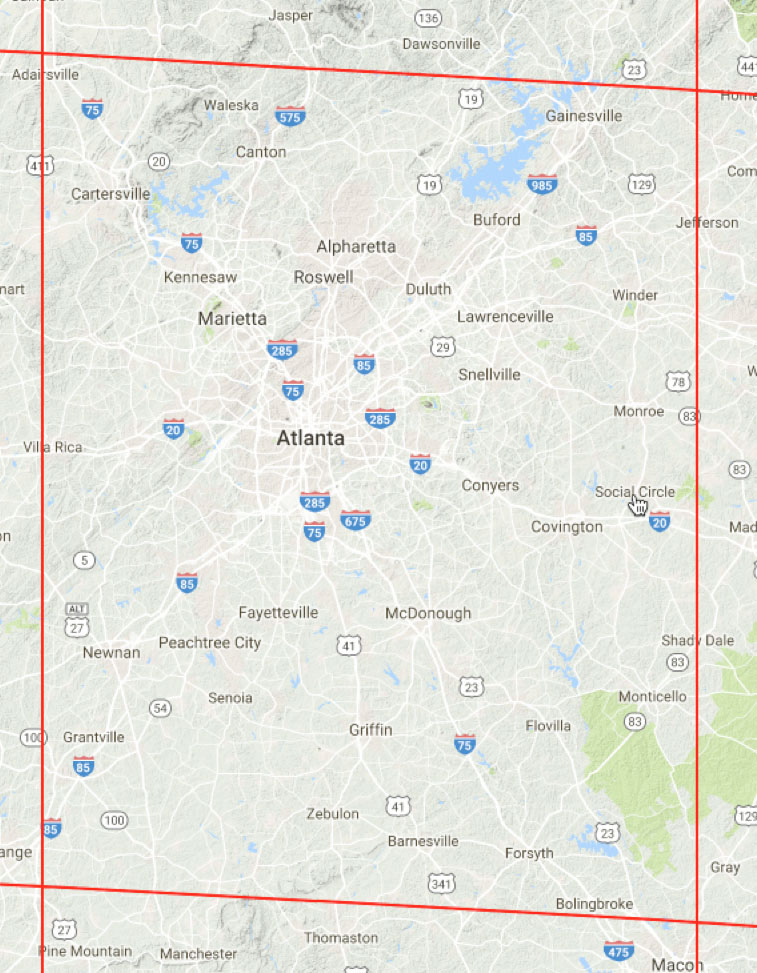} \\
(a) $T_{max} = 100$ & (b) $T_{max} = 200$ & (c) $T_{max} = 250$& (d) $T_{max} = 300$ & (d) $T_{max} = 400$\\
\end{tabular}

\caption{Partitioning of the Atlanta region with minimum S2 level $L_{min} = 6$. The tweet density in this region is lower than in the New York region, thus, reaching the minimum cell level with user threshold of $T_{max} = 400$.}
\label{fig:granularity_atlanta}
\end{figure*}

Table~\ref{table:menet_performance_very_user} shows that when the maximum number of users per cell $T_{max}$ increases, fewer regions are created. The decreasing trend in the mean distance errors  can be explained by the better classification performance when using less classes. Moreover, the median and @$161$ remain stable within the range of $200 - 500$ for $T_{max}$. The reason being that the classification accuracy in this range does not change significantly. The median and the @$161$, however, are much worse with $T_{max}$ set to $100$ even when the corresponding number of regions is limited. The reason being, again, that the classification performance drops dramatically. The question arises: why is the classification accuracy so low? The reason is that  splitting with this setting ignores the geographically natural distribution of data. In fact, an S2 cell at level $6$ is a good fit with the area of cities, where most of the tweets originate. If we lower the user threshold $T_{max}$ in a cell, the splitting algorithm will stop at much higher cell levels for cities where the tweet density is high, thus dividing the city area into multiple smaller regions. That explains why the classification performance is very low. Figures~\ref{fig:granularity_newyork} and ~\ref{fig:granularity_atlanta} show the subvidision of GeoText at level $6$ with different values of $T_{max}$.

\subsubsection{Feature Analysis}
The MENET architecture has the capability of exploiting multiple features according to the multiview learning paradigm. In this paper, we realize the model using four features: TF-IDF, \textit{node2vec}, \textit{doc2vec} and \textit{timestamp}. The question, then, arises: which feature contributes the most to the discriminative strength of the model? 
To answer this question, we conduct additional experiments with different combinations of features.  
Concretely, we eliminate one type of feature from the feature set and perform experiments with the rest. 
This can be done by temporarily removing a branch in MENET just before the concatenation layer (see Fig.~\ref{fig:model1}). For a fair comparison, we use the same parameter setting for MENET as in the experiments with the full feature set. The results from the experiments on the GeoText dataset are presented in Table~\ref{table:dropping_feature_menet}. 

Compared to the results in Table~\ref{table:coordinates_prediction}, it is clear that the  \textit{node2vec} feature is the most important. Removing this feature results in a significant reduction of MENET's performance in terms of mean distance error ($894$ km), median distance error ($480$ km) and accuracy within $161$ km ($36.5$ \%). The contribution of the  \textit{doc2vec} feature is also noticeable, indicating an increase of more than $100$ km in terms of mean distance error, compared to the full feature set. The other features help to improve the performance slightly as removing them results in a marginal decrease in the three performance criteria.

\begin{table}
\centering
\caption{Performance of MENET on GeoText by dropping a feature from the feature set. The experiments are conducted using S2 labels with $L_{min} = 6$ and  $T_{max} = 500$.}
\label{table:dropping_feature_menet}
\begin{tabular}{| c| c c c |}
\hline
Dropped Feature & Mean & Median & @161 \\
& (km) & (km) & (\%) \\
\hline
\hline
TF-IDF & 571 & 35 & 61.4 \\
Node2vec & \textbf{894} & \textbf{480} & \textbf{36.5} \\
Doc2vec & 685 & 65 & 55.4\\
Timestamp & 555 & 33 & 62 \\
\hline
\end{tabular}
\end{table}

\begin{table}
\centering
\caption{Geolocation results on GeoText with $k$-d tree and $k$-means discretization. The experiments are made with $32$ classes.}
\label{table:menet_kdtree_kmeans}
\begin{tabular}{| c | c | c | c | c |}
\hline
Label Type & Mean (km) & Median (km) & @161 (\%) \\
\hline
\hline
$k$-d tree & 573 & 120 & 53.8 \\
$k$-means & 538 & 49 & 61.0 \\
\hline
S2 & \textbf{552} & \textbf{38} & \textbf{62.1} \\
\hline
\end{tabular}
\end{table}

\subsubsection{Performance of MENET with regard to Observed Partitioning}
In Table~\ref{table:coordinates_prediction}, we have a notable improvement in geolocation result with MENET by using S2 labels. Using Google's S2 geometry library is one of many ways to create label sets for our classification problem. A similar partitioning strategy to Google's S2 library is called Hierarchical Equal Area isoLatitude Pixelization of a sphere (HEALPix)~\cite{gorski2005healpix}. Like the S2 library, it is able to partition the sphere into a uniform grid, and has appeared in several papers for Twitter user geolocation such as~\cite{melo2015geocoding}. Other examples include the use of $k$-d tree~\cite{bentley1975multidimensional} and $k$-means~\cite{macqueen1967some} clustering algorithms for grouping users, thus making labels as in~\cite{rahimi2017,rahimi2015}. In this section, we aim at investigating the performance of MENET with respect to two label creation strategies, namely $k$-d tree and $k$-means subdivisions. Our experiments are again conducted on the GeoText dataset. 

Following~\cite{rahimi2017}, we create groups of users using either $k$-d tree or $k$-means partitioning. The clustering of users is based on geographical coordinates, namely latitude and longitude. 
For $k$-d tree subdivision, we make the root node with the bounding box that contains all user coordinates. Then, the tree is made by recursively splitting nodes, which correspond to boxes, into children nodes with straight dividing lines. The splitting takes into account the larger dimension of a node, then tries to divide all users in that node into two groups evenly. Note that we use only leaves to store users, which corresponds to classes. Therefore, the dividing lines must not go through any user's point. The recursive splitting process stops if the number of users in a cell falls below a given threshold. Following~\cite{rahimi2015}, we set the theshold to $300$ resulting in $32$ geographical cells (i.e., classes). When using $k$-means for making classes, the number of clusters is set to $32$,  and the Euclidean distance metric is used. The same hyperparameter settings are kept for MENET in these experiments. 

The geolocation results on the GeoText dataset with \mbox{$k$-d} tree and $k$-means partitionings are shown in Table~\ref{table:menet_kdtree_kmeans}.
It is clear that $k$-means is better than $k$-d tree in partitioning Twitter users, in the sense that it can mitigate the geolocation errors.
Concretely, using $k$-means labels reduces the mean distance error with more than $30$ km. The median distance error reduces by $50$\% while the accuracy within $161$ km improves by roughly $7$\%. 
On the other hand, the performance of MENET using S2 labels is better for all the concerned performance criteria. Also, it is worth mentioning that the performance of MENET with the $k$-means labels is close to that of S2 labels. However, the S2 partitioning is more flexible in controlling the median distance error and it is stable in creating labels compared with $k$-means.

\section{Conclusion and Future Work}
\label{conclusion_future_work}
Noisy and sparse labeled data make the prediction of Twitter user locations a challenging task. 
While plenty approaches have been proposed, no method has attained a very high accuracy. 
Following the multiview learning paradigm,
this paper shows the effectiveness of combining knowledge 
from both user-generated content and network-based relationships.
In particular, we propose a generic neural network model, referred to as MENET, 
that uses words, paragraph semantics, network topology and timestamp information, 
to infer users' location. 
The proposed model provides more accurate results compared to the state of the art, 
and it can be extended to leverage other types of available information, besides the types of data considered in this paper.

The performance of our model heavily depends on user graph features. The \textit{node2vec} algorithm used in this paper is transductive, meaning the graph is built on all users. 
In our future work, we will focus on making the model truly inductive, meaning able to generalize to never seen users. 


%

%



\ifCLASSOPTIONcaptionsoff
  \newpage
\fi

\bibliographystyle{IEEEtran}
\bibliography{references} 

\end{document}